%% file: main.tex
\documentclass[runningheads]{llncs}

\input{packages}

\input{macros}

\usepackage[absolute,overlay]{textpos}

\begin{document}
\title{What, Indeed, is an Achievable Provable Guarantee for Learning-Enabled Safety Critical Systems}
\author{Saddek~Bensalem\inst{1}
\and
Chih-Hong~Cheng\inst{2}
\and
Wei~Huang\inst{3}
\and
Xiaowei~Huang\inst{3}
\and
Changshun~Wu\inst{1}
\and
Xingyu~Zhao\inst{4}
}

\authorrunning{S. Bensalem et al.}
%
\institute{University Grenoble Alpes, VERIMAG, Grenoble, France \\
\email{\{saddek.bensalem,changshun.wu\}@univ-grenoble-alpes.fr }\\
\and
Technical University of Munich, Garching, Germany \\
\email{chih-hong.cheng@tum.de }\\
\and
Department of Computer Science, University of Liverpool, Liverpool, UK \\
\email{\{w.huang23,xiaowei.huang\}@liverpool.ac.uk}
\and
WMG, University of Warwick, Coventry, UK \\
\email{xingyu.zhao@warwick.ac.uk}}

\maketitle  

\begin{abstract}
Machine learning has made remarkable advancements, but confidently utilising learning-enabled components in safety-critical domains still poses challenges.
Among the challenges, it is known that a rigorous, yet practical, way of achieving safety guarantees is one of the most prominent.
In this paper, we first discuss the engineering and research challenges associated with the design and verification of such systems.
Then, based on the observation that existing works cannot actually achieve provable guarantees, we promote a two-step verification method for the ultimate achievement of provable statistical guarantees. 
\end{abstract}

\section{Introduction}
From the studies of Leibniz~\cite{kulstad1997leibniz} to the philosophical view, the human mind and brain have been perceived as an information processing system and thinking as a form of computing. Over three centuries ago, two dreams were mingled, the philosopher's and the engineer's: the philosopher's ideal to have a sound method to reason correctly, and the engineer's dream to have a machine to calculate efficiently and without error. Any attempt to assimilate the human brain into a mechanical or computer machine necessarily negates the autonomy of thought. The latter is not the result of chance or indeterminacy but instead of a possibility of choice according to the reasoning based on rules and principles. By its organization, the human brain allows the emergence of cognitive autonomy. Of course, suppose we accept the idea of a level of existence proper to the cognitive processes. In that case, the philosophical dream becomes, more modestly, that of understanding the diversity of human cognitive functions. The development of the general theory of automata and the formalization of the construction of complex machines by Von Neumann allowed the pursuit of the engineer's dream. A central turning point took place around the 1960s, with the design of machines on the one hand and progress in artificial intelligence (AI) and cognitive science on the other hand. Significant successes have been achieved, for example, in natural language processing.

Our world today is witnessing the genesis of a significant shift in how advanced technologies operate. We are beginning to see increasingly independent and autonomous systems in this emerging wave of available automation. The degree of interactions between these systems and human operators is gradually being reduced and pushed further away. These autonomous systems are inherently sophisticated and operate in complex, unpredictable environments. Unfortunately, they still face deployment concerns in safety-critical applications (e.g., transportation, healthcare, etc.) due to a lack of trust, behavioural uncertainty, and technology compatibility with safe and secure system development methods. In particular, Urban Autonomous Driving and Intelligent Medical Devices are considered to be the most complex problem in autonomy; existing development of autonomous vehicles naturally includes the AI part (e.g., machine-learning for perception), as well as the CPS part (e.g., for vehicle control or decision making via infrastructure support). However, there are significant challenges in ensuring the quality of the overall system.


To ensure the safety of autonomous systems that incorporate AI components, we consider it mandatory for the overall engineering process to understand the safety performance of AI components while considering their impact on the overall system. Guaranteeing safety in critical systems that incorporate AI components, however, is not a straightforward process. Several constituent elements of safety cover all the dimensions of an AI system. The criteria catalog we can find in the literature to improve safety in critical applications can be summarized as the following: 
\begin{itemize}
    \item All algorithms based on decision-making shall be explainable~\cite{gunning2019xai,lapuschkin2019unmasking,confalonieri2021historical,dovsilovic2018explainable};
    \item The functionality of algorithms shall be analyzed and validated using formal verification methods before use~\cite{huang2017safety,dreossi2019verifai,wu2020game,liu2021algorithms,seshia2022toward};
    \item Statistical validation is necessary, mainly in cases where formal verification is unsuitable for specific application scenarios due to scalability issues~\cite{huang2021statistical,zhang2022proa};
    \item The inherent uncertainty of neural network decisions shall also be quantified~\cite{shafaei2018uncertainty,gawlikowski2021survey,hullermeier2021aleatoric,gruber2023sources};
    \item Systems must be observed during operation, for example, by using online monitoring processes~\cite{cheng2019runtime,henzinger2020outside,cheng2020provably,lukina2020into,cheng2022prioritizing}.
\end{itemize}     
In this paper, we promote an approach founded on a two-step integration.
The first 
step involves a system-level analysis and testing, rather than solely focusing on the AI component in isolation.
%
It recognizes the interconnected nature of the system and considers the integration and interactions of various components.
By examining the system as a whole, potential risks and vulnerabilities can be identified, allowing for comprehensive safety assurance.
The second 
step involves a detailed analysis of the AI components themselves, without considering their impact on the overall system.
While this 
step provides insights into the specific AI algorithms and models, 
in itself 
it may overlook potential risks arising from the interactions between the components and the broader system context. 
%
This two-step integration of verification processes is to assess the safety performance of AI components while also considering their impact on the overall system. 
This entails examining not only their individual performance but also their interactions within the broader system context.
In addition to formal analysis, we can also conduct studies on statistical guarantees and how these guarantees propagate throughout the system.

The rest of the paper is organized as follows.
Sections 2 and 3 discuss the challenges of designing reliable and trustworthy AI critical systems from an engineering and research perspective, respectively.
In Section 4, we present our methodology and proposed solutions to tackle these challenges. 
Finally, Section 5 provides a summary of the conclusions and highlights avenues for future work.

%

\section{Challenges in Engineering Safety-critical Systems integrating Learning-enabled Components}\label{sec:gaps}

The engineering of safety-critical systems has been a mature paradigm with the support of safety standards such as IEC 61508, ISO 26262, or DO-178c. The rigorous method implied by the process focuses on  hazards caused by the malfunctioning behavior of E/E safety-related systems, including the interaction of these systems. Nevertheless, even in the absence of system malfunctioning, functional insufficiencies caused by performance limitations and incomplete/improper specification can also be the source of hazards, where standards such as ISO~21448 are introduced to address these issues. 

To ensure the necessary level of safety and reliability, a learning-enabled component must also meet the identical functional safety standards encompassing reliability, applicability, maintenance, and safety (RAMS) as any other system. Moreover, it should mitigate the impacts of malfunctions to fulfill the essential safety and reliability prerequisites. On the other hand, properly ensuring the safety of the intended functionality (SOTIF) is the crucial gap in embracing the legitimate use of learning-enabled components. In the following, we enumerate some of the key limiting factors. 

\begin{enumerate}
    \item The introduction of learning-enabled components comes with the practical motivation where the operational environment is \textbf{\emph{open}} and \textbf{\emph{dynamic}} (e.g., urban autonomous driving), thereby inherently making systematic analysis complicated. 

    \item \textbf{\emph{Data}} has played a central role in learning-enabled systems. Under the slogan  ``data is the new specification'', it is crucial to have a systematic approach to performing data collection, cleaning, labeling, as well as managing the data to incorporate adjustment of the operational domain.   

    \item Learning implies translating the implicit knowledge embedded in the data to a model. Despite the mathematical optimization nature of learning model parameters being transparent, the \textbf{\emph{uncertainty}} caused by the model training or the data can lead to fundamental concerns about the validity of the prediction. 

    \item Classical techniques for software \textbf{\emph{verification}}  encounter scalability   issues. Learning models such as deep neural networks create highly non-linear functions to perform classification and prediction tasks. Formal verification or bug finding thus can be viewed as a non-linear optimization problem across the high dimensional input space. The problem even worsens when the learned model controls a \textbf{\emph{plant}} governed by  highly nonlinear dynamics. 

    \item The derivation of \textbf{\emph{safety specifications for the learning component}} can be far from trivial. While for tasks such as image-based object detection, the performance specification characterizing the error rate is relatively straightforward (which commonly leads to a probabilistic threshold on error rate), for control applications, the safety and performance requirement needs to be translated into reward signals in order to be used by (reinforcement) learning methods.

    \item Finally, the above challenges are further complicated by the fact that the engineering of learning-enabled components is \emph{\textbf{iterative}} with the goal of \emph{\textbf{continuous improvement}}. It is also complicated by the dimension of avoiding malfunctioning, implying the need to design hardware or software architectures to avoid transient or permanent faults in the learning-enabled components. 
\end{enumerate}

Unfortunately, the state-of-the-art guidelines or standards only provide high-level principles, while concrete methods for \textbf{\emph{safe and cost-effective}} implementation are left for interpretation. This ultimately brings the research need in the field, which we detail in subsequent sections.

\section{Research Challenges}
%
%

The reason why one wants to apply machine learning to a safety critical application is two-fold: (1) it is impossible to program a certain functionality of the application and (2) a machine learning model can not only perform well on existing data but also generalise well to unseen data. Nevertheless, it is required that a machine learning model has to be safe and well performed such that both safety and performance can be quantified with error bounds given. Safety will be prioritised when a balance is needed. 

\begin{remark}\label{remark:challenges}
    While non-trivial, it is possible that a software or hardware system can be designed and implemented with ultra-high reliability, thanks to the availability of specification and requirements. However, this is unlikely for machine learning models, due to the unavailability of specifications and the complexity of the learning process. This calls for novel design and implementation methodologies for machine learning systems to satisfy both safety and performance requirements. 
\end{remark}

For the remaining of this section, we discuss challenges a novel methodology needs to tackle.
While every gap between traditional software and AI-based systems, as discussed in Section~\ref{sec:gaps}, leads to research challenges, we believe the most significant ones are from (1) the environmental uncertainties that an AI-based system has to face, (2) the size and complexity of the AI models themselves, and (3) the lack of novel analysis methods that are both rigorous and efficient in dealing with the new problems. These three challenges lead to our proposal of considering \textbf{\emph{statistical guarantees}} and \textbf{\emph{symbolic analysis}} of AI models.

\subsection{Uncertainty}

In machine learning, uncertainty is often decomposed into aleatoric uncertainty and epistemic uncertainty, with the former irreducible and the latter reducible in theory. To explain this, we formalise the concept of generalisability, which requires that a neural network  works well on all possible inputs in the data domain $\inputdomain$, although it is only trained on the training dataset $(X,Y)$. 


\begin{definition} 
	Assume that there is a ground truth function $f: \inputdomain\rightarrow \outputdomain$ and a probability function $O_p: \inputdomain\rightarrow [0,1]$ representing the operational profile. A network $\network$ trained on $(X,Y)$ has a generalisation error: 
	\begin{equation}
G^{0-1}_\network = \sum_{x\in \inputdomain}  {\bf 1}_{\{\network(x) \neq f(x)\}}  \times O_p(x)
\label{eq_gen_error_01}
\end{equation}
where ${\bf 1}_{\tt S}$ is an indicator function -- it is equal to 1 when {\tt S} is true and 0 otherwise.
\end{definition}

We use the notation $O_p(x)$ to represent the probability of an
input $x$ being selected, which aligns with the \textit{operational profile} notion \cite{musa_operational_1993} in software engineering. 
Moreover, we use 0-1 loss function (i.e., assigns value 0 to loss for a correct classification and 1 for an incorrect classification) so that, for a given $O_p$, $G^{0-1}_\network$ is equivalent to the reliability measure \textit{pfd} (the expected probability of the system failing on a random demand) defined in the safety standard IEC-61508. 

We decompose the generalisation error into three: 
\begin{equation}
\label{eq_decomp_ge}
G^{0-1}_\network =\underbrace{G^{0-1}_{\network} -\inf_{\network \in \networks}G^{0-1}_\network}_\text{Estimation error of $\network$}
+
\underbrace{\inf_{\network \in \networks}G^{0-1}_\network-G^{0-1,*}_{f,(X,Y)}}_\text{Approximation error of $\networks$}
+\underbrace{G^{0-1,*}_{f,(X,Y)}}_\text{Bayes error}
\end{equation}

\textit{a)} The \textit{Estimation error of $\network$} measures how far the learned classifier $\network$
is from the best classifier in $\networks$,  the set of possible neural networks with the same architecture  but different weights with $\network$. Lifecycle activities at the \textbf{model training} stage 
essentially aim to reduce this error, i.e., 
performing
optimisations 
of 
the set $\networks$.

\textit{b)} The \textit{Approximation error of $\networks$} measures how far the best classifier in $\networks$ is from the overall optimal classifier, after isolating the Bayes error. The set $\networks$ is determined by the architecture of DNNs (e.g., numbers of layers
), thus lifecycle activities at the \textbf{model construction} stage are used to minimise this error.


\textit{c)} The \textit{Bayes error} is the lowest and irreducible error rate over all possible classifiers for the given classification problem \cite{fukunaga_introduction_2013}. The irreducibility refers to the training process, and the Bayes errors can be reduced in data collection and preparation.  It is non-zero if the true labels are not deterministic (e.g., an image being labelled as $y_1$ by one person but as $y_2$ by others), thus intuitively it captures the uncertainties in the dataset $(X,Y)$ and true distribution $f$ when aiming to solve a real-world problem with machine learning. We estimate 
this error
(implicitly) 
at the \textbf{initiation} and \textbf{data collection} stages in activities like: necessity consideration and dataset preparation etc.


Both the Approximation and Estimation errors are reducible, and are caused by the epistemic uncertainties. The Bayes error is irreducible, and caused by the aleatoric uncertainty. The \emph{ultimate goal} of all lifecycle activities is to reduce the three errors to 0,  
especially for safety-critical applications. 


\textbf{Aleatoric uncertainty}, as discussed in \cite{gruber2023sources}, may include various data-related issues such as survey error, missing data, and possible shifts in the data when deployed in real-world. Definition and measurement of these uncertainties can be done more reasonably with probabilistic/statistical distributions. 


\subsection{Size and Complexity of the AI Models}


Another significant challenge is on the AI model itself. While it is believed that larger -- often overparamterised --  models can perform well \cite{Nakkiran2020Deep}, large models cannot be analysed analytically due to its size and complexity. Formal verification methods that check the local robustness of a neural network against perturbations are either limited on the number of neurons in a network (such as \cite{10.1007/978-3-030-32304-2_15,10.1145/3368089.3417918,10.1007/s00165-021-00548-1}) or limited by the number of input dimensions that can be perturbed (such as \cite{ruan_reachability_2018,ruan_global_2019,peipeiCIS2022}). Even the theoretical analysis in machine learning field, which is usually less rigorous than in formal methods, has to be conducted on much simpler models such as linear and random projection models (see e.g., \cite{doi:10.1073/pnas.1903070116}). The situation is getting worse when we have to deal with large language models, see e.g., \cite{huang2023survey} for a discussion on their safety and trustworthiness issues. 

\subsection{Lack of Novel Analysis Methods that are both Rigorous and Efficient}

Existing analysis methods from either formal methods or software testing are mostly aimed to extend their success in traditional software and systems. For example, constraint solving and abstract interpretation are popular methods in robustness verification of neural networks, and structural testing are popular testing methods for neural networks. 
Actually, the traditional verification methods have already been experiencing scalability problems when dealing with traditional software (up to several thousands lines of code), and it is therefore unlikely that they are able to scale and work with modern neural networks (which typically have multi-millions or billions of neurons). For testing methods, there is a methodological barrier to cross because neurons do not have clear semantics as variables, so does the layers with respect to the statements. Such mismatches render the test coverage metrics, which are designed by adapting the known test coverage in software testing such as statement coverage, potentially uncorrelated with the properties to be tested. 

Another critical difference from traditional software that is posed on the analysis methods is the perfection of neural networks. For software to be applied to safety critical applications, a ``possible perfection'' notion \cite{littlewood_reasoning_2012,rushby_software_2009,zhao2017modeling} is used. However, for machine learning, the failures are too easy to find, and it does not seem likely that a perfect, or possibly perfect, machine learning model exists for a real-world application. To this end, a novel design method is  needed to ensure that an AI-based system can potentially be free from serious failures.   

Moreover, multiple properties may be required for a machine learning model, e.g., robustness, generalisation, privacy, fairness, free from backdoor attacks, etc. However, these properties can be conflicting (e.g., robustness-accuracy trade-off) and many of them without formal specifications, which lead to the challenge of lacking effective methods for the analysis and improvement of them altogether for a machine learning model. 


\section{Methodology}

The needs of safety critical systems require that,  even facing challenges that are more significant than  traditional software, a legitimate methodology will still provide rigorous and provable guarantees, concerning the satisfiability of properties on the autonomous cyber-physical system under investigation. 
{We conceptualise AI-based systems into five levels (shown in Fig.~\ref{fig_levels_rq}). For the remainder of this section, we discuss the methodology needed at each level and across levels. {Specifically, at each level, we consider the following questions: For sources of uncertainty identified in earlier sections, what metrics (e.g., binary, worst-case or probabilistic) shall we use to measure them? How to efficiently evaluate those metrics? Can we provide any forms of guarantees on the evaluations? Moreover, we raise questions that span across different levels: How do metrics at higher levels break down to metrics at lower levels? If and how the guarantees (in various forms) from lower levels can propagate and compound to higher levels, ultimately aiming to make meaningful claims about the entire system.} 
\begin{figure}[http]
    \centering
    \includegraphics[width=0.9\textwidth]{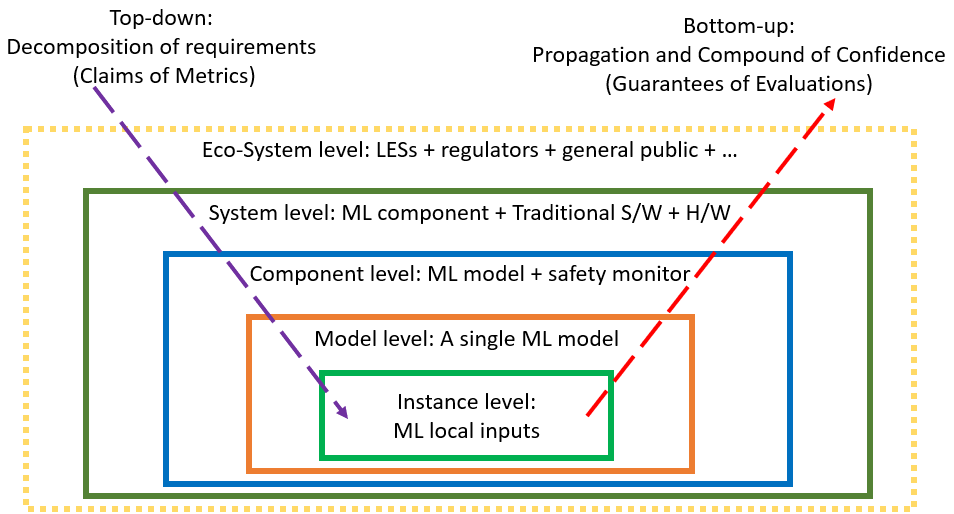}
    \caption{Research challenges organised into five conceptual levels with top-down and bottom-up routes.}
    \label{fig_levels_rq}
\end{figure}

Our proposed methodology consists of the following attributes: 
\begin{itemize}
    \item a set of specification languages that describe, and connect, requirements of different levels; 
    \item a formal method of propagating statistical guarantees about the satisfiability of the requirements across the system and the component levels; 
    \item a rigorous method of achieving required statistical guarantees at the instance and the model levels; 
    
\end{itemize}

This is founded on two threads of state-of-the-art research: a design and co-simulation framework (Section~\ref{sec:designflow}) and some design-time verification and validation (V\&V) methods for machine learning models  (Section~\ref{sec:vnvframework}). 
The co-simulation framework is 
to effectively simulate the real-world system and environment to a sufficient level of fidelity. The V\&V methods are to detect vulnerabilities and improve the machine learning model's safety and performance. 
It is noted that, the  V\&V methods can improve the system but may not be able to provide provable safety guarantee, for reasons that we will discuss below. Once the improvement is converged (or certain termination condition is satisfied), the new methodology is applied for the ultimate achievement of provable guarantees.


\subsection{State-of-the-Art 1: A Design and Co-Simulation Framework}\label{sec:designflow}

This section summarizes the current research on the rigorous design of AI-enabled systems out of the EU Horizon 2020 project FOCETA. 


\subsubsection{Design of trustable AI models}
Design of trust-able AI models requires considering the complete engineering life-cycle beyond optimizing the model parameters.
Figure~\ref{fig:lifecycleVVMethods} presents a flow design for the AI model development. We consider the lifecycle phases: data preparation, training, offline verification and validation, and online deployment. During the offline V\&V, techniques for the falsification and explanation are applied to discover whether there are failures regarding the decision-making (i.e., falsification) or failures demonstrating the inconsistency with human's perception (i.e., explanation). In addition to their individual functionalities, falsification and explanation may benefit from mutual interactions, 
to make sure that a decision failure can be explained and two inconsistent explanations are tested, see e.g., \cite{huang2022safari}. A formal verification process is called only when no error can be found from both falsification and explanation. 


\begin{figure}[http]
    \centering
    \includegraphics[width=0.95\textwidth]{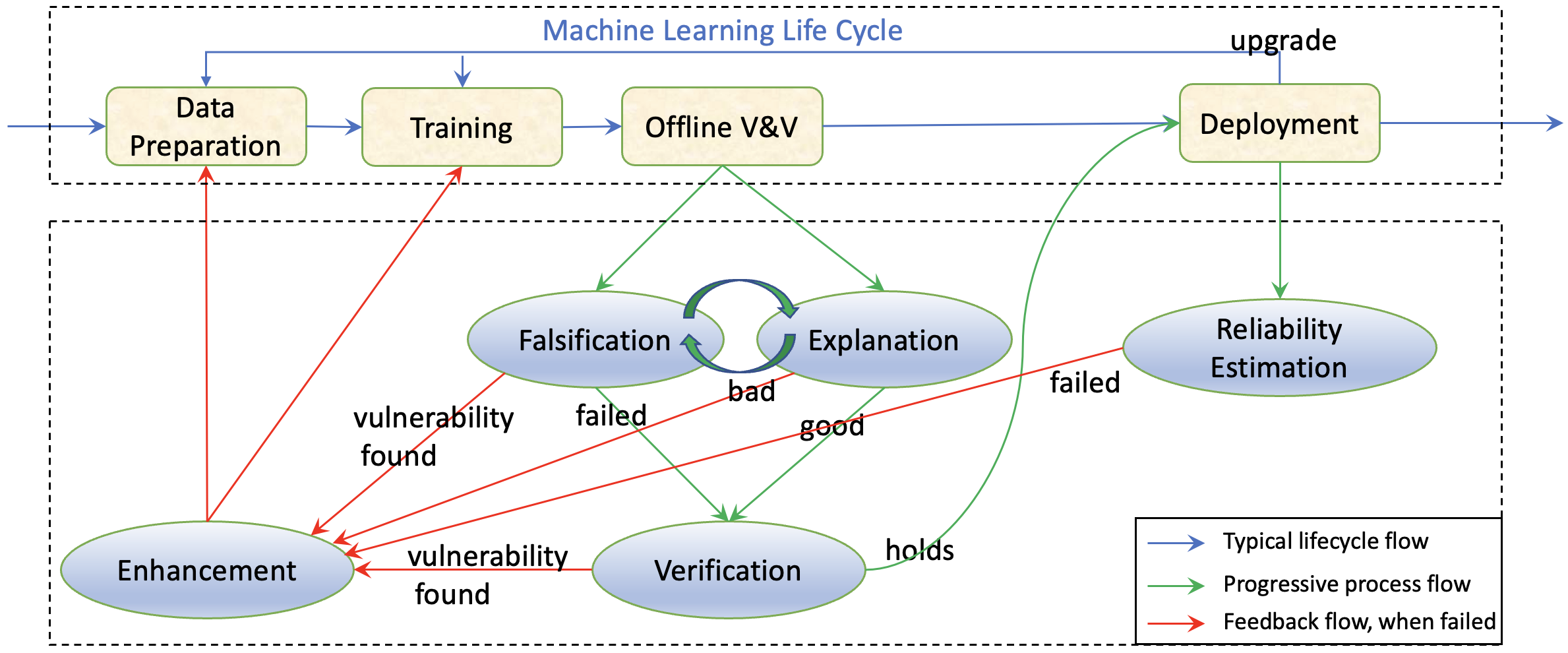}
    \caption{A Verification and Validation Framework for Machine Learning Enhancement}
    \label{fig:lifecycleVVMethods}
\end{figure}

In the context of a real-world learning-enabled system, the offline V\&V can be insufficient, due to the scalability of the verification techniques and the  environmental uncertainties that are unknown during offline development (details will be provided below). In such cases, 
a reliability estimation to analyse the recorded runtime data will be needed, to understand statistically whether the AI-based system can run without failures e.g., in the next hour, with high probability. 

Another important module in Figure~\ref{fig:lifecycleVVMethods} is the enhancement, where the failure cases are considered for the improvement of the machine learning models, through either data synthesis or model training. 


\subsubsection{Design flow for safety-critical systems with AI components}
Like any other safety-critical systems, the design flow for AI-enabled systems shall also cover design and operation time activities.
%
%
However, in contrast to classical critical systems where the environment is largely static and predictable, the use of AI-enabled systems reflects the need to handle an open environment. 

Within FOCETA, we view the engineering of the complete system as analogous to the engineering of the AI component, where it is important to create a continuous loop of improvement between development and operation. The current state of the practice is extended toward transferring knowledge about systems and their contexts (e.g., traffic situations) from the development to operations and from the operations back to action in iterative steps of continuous improvements. The methodology enables their ongoing engineering over the complete life cycle of autonomous learning-enabled systems -- from specification, design, implementation, and verification to operation in the real world with a particular focus on correctness concerning evolving requirements and the systems’ safety. Moreover, the whole design flow ensures traceability between requirements and the system/component design.


A key feature is the usage of runtime monitors for the seamless integration of development and operations. In contrast to AI component monitors that largely detect situations such as out-of-distribution, system-level runtime monitors observe a system (part) via defined interfaces and evaluate predefined conditions and invariants about the system behavior based on data from these interfaces. This allows us to identify the need for AI model updates during continuous testing/verification if, for example, some data in a test scenario results in a safety property violation or if a new requirement emerges in response to a previously unknown adversarial threat.

\subsubsection{Simulation-based Modeling and Testing at Design Time}
Using simulation in the design phase offers multiple advantages.
It provides a cost-effective means to verify the system’s performance over diverse parameter ranges and generate massive scenarios.
It follows that critical methods can be already identified in the virtual environment, and only those can be replayed in the much more expensive physical setting.
Simulation-based testing allows the generation of scenarios (e.g., with infrequent events) that may be impossible to realize when the system operates in its environment (e.g., with specific weather conditions, such as fog or snow).
It also enables the creation of safety-critical situations (e.g., crashes) without compromising the safety of the real-life actors.
%
%
%
%
The simulation framework shall allow the integration of heterogeneous components that may be designed with different tools and frameworks. In addition, there is a need to rigorously argue that the domain gap between synthetic data produced by the simulation engine and real data observed in the field is closed. In layman's words, an image being ``photo-realistic" does not necessarily imply its being ``real".  
%

\subsubsection{Deployment, Operation, and Analysis of AI critical systems at Runtime}
The analysis of the AI components and their integration into the AI critical system during design time, together with protective mechanisms synthesized around AI models, help the safety assurance of the overall system during real-time operation.
These measures are complemented by runtime verification, which plays a central role during the AI critical operation.
Runtime monitors allow us to observe the system and its interaction with the environment and gather helpful information regarding (1) violation of safety or other requirements and (2) new operational scenarios that were not captured by the training data and models, and (3) other unexpected situations and anomalies not characterized by the existing requirements set. 
To be effective, monitors must be present both at the component level (AI and classical) and at the system level; the information gathered by different monitors must be fused to derive useful information that can be used to i) ignore the situation (e.g., detected object misclassification) that does not impact the system-level control decision, ii) take a protective measure (e.g., switch from the advanced to a base controller) or iii) improve the design (e.g., provide a new scenario for the training data).
The last point refers to an evolvable AI critical system, in which information from the system operation is collected and used to go back to the design and enhance its functionality based on new insights, thus effectively closing a loop between the design and the operation phase. 

\subsection{Properties and Specifications at Different Levels}

On the system level, we may use temporal logic to express the required dynamic behavior. There are recent attempts to extend the temporal logic for AI-based systems. For example, \cite{FRET} formalises requirements 
of an autonomous unmanned aircraft system based on an extension of propositional LTL, where temporal operators are augmented with timing constraints. {It  uses atomic 
propositions such as ``$horizontalIntruderDistance>250$'' to express the result of the perception module,  without considering the sensory input and the possible failure of getting the exact value for 
$horizontalIntruderDistance$.} 
\cite{BCHKMNP2022} introduces a specification language based on LTL, which utilises event-based abstraction to hide the details of the neural network structure and parameters. {It considers the potential failure of the perception component and  uses a predicate $pedestrian(\textbf{x})$ to express if $\textbf{x}$ is a pedestrian (referring to the ground truth). However, it does not consider the predicate's potential vulnerabilities,  such as robustness,  uncertainty, and backdoors.}  \cite{BPQ} proposes Timed Quality Temporal Logic (TQTL) to express monitorable~\cite{10.1007/978-3-030-88494-9_18} spatio-temporal quality properties of perception systems based on neural networks. {It considers object detectors such as YOLO and uses expressions such as $D_0 : d_1: (ID, 1), (class, car), (pr, 0.9), (bb, B1)$ to denote an object $d_1$ in a frame $D_0$ such that it has an index $1$, a predictive label $car$, the prediction probability $0.9$, and is in a bounding box $B_1$. Therefore, every state may include multiple such expressions, and then a TQTL formula can be written by referring to the components of the expressions in a state. }

For our purpose of having a statistical guarantee for properties at the system level (see Figure~\ref{fig_levels_rq}), for any temporal logic formula $\varphi$, a statistical guarantee is needed, e.g., in the form of 
\begin{equation}\label{equ:pacguarantee}
  P(err(\varphi)\leq \epsilon) > 1-\delta
\end{equation}
where $\varphi$ is a formula such that $err(\varphi)$ denotes that estimation error on the satisfiability of $\varphi$ on the system,
and $\epsilon$ and $\delta$ are small positive constants.
In the formula $\varphi$, we need 
atomic propositions that are related to the perception components. According to different assumptions, we may have different atomic propositions: instance-level atomic propositions or model-level atomic propositions. 
For an instance-level atomic proposition such as $pedestrian_{\epsilon,\delta}$, it expresses that the error of having a $pedestrian$ in the current input is lower than $\epsilon$, under the confidence level no less than $\delta$. In such cases, the statistical guarantee is established by considering the local robustness (i.e., cell unastuteness as in \cite{dong2023reliability}). On the other hand, for a model-level atomic proposition such as $perception_{\epsilon,\delta}$, it expresses that the error of having a failed detection among all possible next inputs is lower than $\epsilon$, under the confidence level no less than $\delta$. In such cases, the statistical guarantee is established by considering the reliability as in \cite{dong2023reliability}. 

The selection between instance-level and model-level atomic propositions depends on the assumptions. If we believe that a failure on the perception component does not have a correlation with the failures of other components, a model-level atomic proposition will be sufficient. On the other hand, if a correlation between failures is expected, and we want the verification to fully consider such correlations, an instance-level atomic proposition will be more appropriate and accurate. 


Section~\ref{sec:guaranteecomponent} will discuss how to achieve the statistical guarantee (i.e., $\epsilon$ and $\delta$). For the model-level atomic propositions, the specification language in \cite{10.1007/978-3-031-17244-1_1} considers not only the functionality (i.e., the relation between input and output) of a trained model but also the training process (where objects such as training datasets, model parameters, and distance between posterior distributions are considered). With this, it can express the safety and security properties that describe the attacks during the lifecycle stages.

\subsection{Guarantees Achieved at Component Levels}\label{sec:guaranteecomponent}
%

This section will discuss a potential solution that can be utilised to achieve the statistical guarantee (i.e., $\epsilon$ and $\delta$) for an atomic proposition describing certain safety properties as summarised in \cite{10.1007/978-3-031-17244-1_1}.  As discussed in Section~\ref{sec:vnvframework}, this cannot be achieved by a standalone machine learning model, even if a V\&V framework as in Figure~\ref{fig:lifecycleVVMethods} is applied, due to the insufficiency of machine learning models. 
We suggest a monitored machine learning system, i.e., a machine learning model running in parallel with a runtime monitor. As indicated in Figure~\ref{fig:runtime}, for non-ML safety-critical systems with clear specifications about safety, a runtime monitor often acts like an alarm to alert the unsafe behaviour. 
On the other hand, for ML systems without safety specifications but only with samples, a runtime monitor needs to analyze the samples and predict the safety of the current input e.g., in the manner of a traffic light system as briefly discussed below. While the predictions might not be completely correct, we expect they are conservative with a provable guarantee and as accurate as possible. 
\begin{figure}[http]
    \centering
    \includegraphics[width=0.8\textwidth]{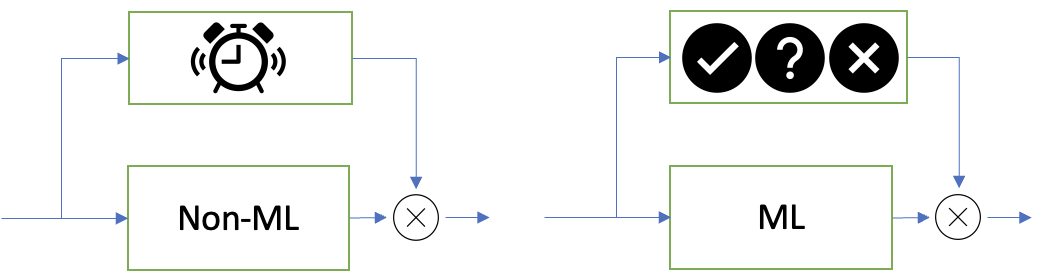}
    \caption{Runtime Monitors with (for non-ML systems) and without (for ML systems) specifications}
    \label{fig:runtime}
\end{figure}

A runtime monitor checks every input of a neural network and issues warnings whenever there is a risk that the neural network might make wrong decision. 
As discussed earlier, given the availability of many adversarial attacks, it is unlikely that a neural network itself can achieve ``possible perfection'' \cite{zhao2020safety} -- a notion of traditional safety-critical software introduced by \cite{littlewood_reasoning_2012,rushby_software_2009,zhao2017modeling}. The safety of a neural network, however, can potentially be achievable with the support of a runtime monitor. Actually, in the extreme case, if a runtime monitor is so restrictive that none of the input instances can pass without warning, the neural network under the runtime monitor is safe (although the performance is bad). 

The design of a runtime monitor for a given neural network is to ensure that the safety of the monitored neural network can be achieved with guarantees. While the restrictive runtime monitor mentioned above suggests absolute safety, it is also undesirable due to its performance. In the following, we discuss a possible runtime monitor that is able to achieve statistical guarantee of the form (\ref{equ:pacguarantee}). 
The core idea of this monitor is to represent the abstracted experiences symbolically, serving as references for future behaviors.
The process involves recording observed data or their learned high-level features for each decision made by the neural network.
These data points are then clustered based on their similarities, and each cluster is approximated by a box as an abstraction. 
Every box $\textbf{b}$ can be described as a tuple $(l,r,c,m,y,i)$, where $l$ is the location of the box, $r$ is the radius vector of the box, $c$ is the cluster that the box belongs to, $m$ is the number of data samples that the box contains, and $y$ is the predictive class label of the box, and $i$ is the correctness indicator of prediction relating to the abstracted samples. 
Once these abstractions are derived, they can be effectively symbolized, and operational symbols can be defined to establish the runtime monitor.
When new data points and decisions arise, we compare the network's behavior for a given input with the reference abstractions. 
Generally, there exist two types of boxes: \textit{positive} and \textit{negative} ones, representing abstracted good and bad behaviors, respectively. 
If the behavior is similar to good behaviors (inside a positive box), the decision is accepted; if it resembles bad behaviors (inside a bad box), the decision is rejected.

Our guarantees are on two levels. 
The first level considers the confidence the runtime monitor classifies an input as safe or not. 
Assume that we have a data point that falls within a box $\textbf{b}$ that is either positive or negative with respect to a label. Since the box will classify the data point, we can utilise the information in the box (e.g., the known points that fall within the box) to conduct a Probably Approximately Correct (PAC) analysis, or utilising Hoeffding inequality such as in \cite{huang2021statistical}, to determine a probability and an error. That is, for each box $\textbf{b}$, we may have 
\begin{equation}\label{equ:firstlevel}
    P_{\textbf{b},\mathcal{D}}(err_{\textbf{b},\mathcal{D}}(c,h)\leq \epsilon_{\textbf{b}}) > 1-\delta_{\textbf{b}}
\end{equation}
for small positive numbers $0< \epsilon_{\textbf{b}}, \delta_{\textbf{b}} <1/2$, where $\mathcal{D}$ is the data distribution, $err_{\textbf{b},\mathcal{D}}(c,h)$ is the probabilistic error of the hypothesis $h$ (i.e., the probability of $h$ does not hold) 
in the box $\textbf{b}$, with respect to the concept $c$ and the distribution $\mathcal{D}$, such that $err_{\textbf{b},\mathcal{D}}(c,h) = P_{x\in \mathcal{D}, x\in \textbf{b}}(h(x)\neq c(x))$, i.e., the probability over $x$ drawn from $\mathcal{D}$ and $\textbf{b}$ that $h(x)$  and $c(x)$ differ.

The second level considers the probability of a future input that our runtime monitor can confidently classify. Assume that we have a set of $n$ boxes in the space (e.g., the hidden space before the Softmax layer) such that they are either positive or negative (we do not consider uncertain boxes for guarantees) with respect to certain label. We can use hypothesis testing to determine the probability (the error) of a future point falling within these boxes, according to the set of known data points. 
Similarly, we will have 
\begin{equation}\label{equ:secondlevel}
    P_{\mathcal{D}}(err_{M^*,\mathcal{D}}(M)\leq \epsilon_{M^*}) > 1-\delta_{M^*}
\end{equation}
for small positive numbers $0< \epsilon_{M^*}, \delta_{M^*} <1/2$, where $err_{M^*,\mathcal{D}}(M) = P_{x\in \mathcal{D}}(M(x)\neq M^*(x))$ is the probability that the runtime monitor $M$ disagrees with the ground truth $M^*$ regarding whether an input $\textbf{x}$ is within the confirmed boxes. 
Moreover, we can replace the hypothesis testing with more effective, and more scalable, probability methods such as MCMC or that we did in \cite{zhao2020safety}.

A ``combination'' (to be analytically derived as future work) of the above levels will reach a statistical way of conducting reliability estimation over runtime data.
A statistical guarantee of the form  
\begin{equation}\label{equ:componentguarantee}
    P_{\mathcal{D}}(err_{\mathcal{D}}(c,h)\leq \epsilon) > 1-\delta
\end{equation}
will be achieved, where $\mathcal{D}$ is the operational distribution of the AI component within the system, $err_{\mathcal{D}}(c,h)$ denotes the error probability of the AI model $c$ with respect to the ground truth $h$, and both $\epsilon$ and $\delta$ are small positive numbers. 

\begin{remark}
    The chance constraint (as in Equation (\ref{equ:componentguarantee})) as a statistical guarantee for safety is not as strong as a deterministic guarantee, which states the absolute missing of failures, or a probabilistic guarantee, which states the missing of failures with certain probability. However, the deterministic guarantee is infeasible in practice due to the environmental uncertainties, as we have discussed for offline verification and validation. Practical methods are missing on how to achieve tight probabilistic guarantees.  
\end{remark}

\subsection{State-Of-The-Art 2: Offline V\&V Methods and Guarantee}\label{sec:vnvframework}



This section discusses the existing verification and validation methods, and explains why they cannot provide the guarantees that are needed for AI-based systems. 
AI models, especially Deep Neural Networks, are known to be susceptible to the adversarial attack and backdoor attack. Given a DNN model $f$, which maps a high dimensional input $x$ to a prediction class $y$, adversarial attack and backdoor attack add maliciously generated perturbations $\epsilon$ into the benign inputs, leading to the mis-predictions of DNNs (refer to the survey for the difference between adversarial attack and backdoor attack).
\begin{equation}
    f(x) = y \;\&\; f(x+\epsilon)\neq y
\end{equation}
This section will briefly review the existing V\&V methods on the robustness of DNNs against the adversarial perturbation $\epsilon$ and discuss the guarantee to the safety of AI model.

\paragraph{Verification} Verification techniques are to determine whether or not a property of a neural network holds within a given range of inputs. The existing verification techniques can be categorized according to the guarantee they provide. \textit{Deterministic guarantees} are achieved by transforming verification of deep neural networks into a set of constraints so that they can be solved with a constraint solver, such as Satisfiability Modulo Theories (SMT) solver~\cite{katz2017reluplex,ehlers2017formal},  Boolean Satisfiability Problem (SAT) solver~\cite{narodytska2018formal,narodytska2018verifying,cheng2018verification}, and mixed integer linear programming (MILP) solver~\cite{cheng2017maximum,lomuscio2017approach}. The name “deterministic” comes from the fact that these solvers often return a deterministic answer to a query, i.e., either satisfiable or unsatisfiable. Some verification techniques can offer \textit{one-sided guarantee}, i.e. deep neural network is robust when adversarial perturbation measured by $L_p$ norm is bounded less than $\epsilon$. These approaches leverage the abstract interpretation~\cite{gehr2018ai2,mirman2018differentiable}, convex optimization~\cite{wong2018provable,dvijotham2018dual}, or interval arithmetic~\cite{wang2018formal,peck2017lower} to compute the approximate bound. Compared to the verification techniques with deterministic guarantee, the bounded estimation can work with larger models, up to 10000 hidden neurons, and can avoid floating point issues in existing constraint solver implementations ~\cite{neumaier2004safe}. To deal with real-world system, which contains the state of the art DNNs with at least multi-million hidden neurons, some practical verification techniques are developed to offer \textit{converging bounds guarantee} and \textit{statistical guarantee}. Layer-by-layer refinement~\cite{huang2017safety}, reduction to a two-player turn-based game~\cite{wu2020game}, and global optimization-based approaches~\cite{ruan2018reachability} are developed to compute the lower bounds of robustness by utilizing the Lipschitz constant and the bounds converge to the optimal value. The \textit{statistical guarantee} is achieved by utilizing the statistical sampling methods, e.g. Monte Carlo based sampling, to estimate the robustness with a certain probability. CLEVER~\cite{weng2018evaluating} estimates the robustness lower bound by sampling the norm of gradients and fitting a limit distribution using extreme value theory. \cite{webbstatistical} utilizes the multi-level splitting sampling to calculate the probability of adversarial examples in the local region as an estimation of local robustness. The local probabilistic robustness estimation can be aggregated over the train set to form the global robustness estimation~\cite{wang2021statistically}. \cite{dong2023reliability,zhao2021assessing} further propose the concept of reliability, which is a combination of robustness and generalization, and estimated on the operational dataset to provide statistical guarantee on neural networks' overall performance.

\paragraph{Testing} When working with large-scale models, often used in the industry, verification is not a good option. Verification techniques offer guarantees to the results at the expense of high computational cost. The cost goes sharply with the increase of model's complexity. Testing arises as a complement to verification. Instead of pursuing mathematics proofs, testing techniques exploit the model in a broad way to find potential faults. The first category of works is the coverage-guided testing. A large amount of coverage metrics are designed in consideration of the structure information of DNNs. Structure coverage metrics, such as neuron coverage~\cite{pei2017deepxplore}, k-multisection neuron coverage~\cite{ma2018deepgauge}, neuron activation pattern coverage~\cite{ma2018deepgauge}, Modified Condition/Decision Coverage (MC/DC) for neuron layers~\cite{sun2018concolicb} are proposed in the past few years. There are also a few works dedicated to designing coverage metrics for Recurrent Neural Networks (RNNs), such as modeling RNNs as abstract state transition systems and covering different states and transitions~\cite{du2019deepstellar}, and quantifying one-step hidden memory change and multi-step temporal relation~\cite{huang2021coverage}. They are all based on the assumption that the activation of neurons represents the functionality of DNNs. By achieving a higher coverage rate in proposed structure coverage metrics, the functionality of DNNs are more thoroughly exercised. Therefore, structure coverage metrics can guide the generation of test cases as diversified as possible, and detect different types of defects, such as adversarial examples and backdoor input~\cite{huang2021coverage}. However, the weak correlation between structure coverage metrics and the defects can not guarantee that increasing the coverage rate can find more faults in DNNs.

The second category of works is distribution-aware testing. There has been a growing body of research focusing on the development of distribution-aware testing techniques for DNNs. To approximate the distribution of training data, deep generative models such as Variational AutoEncoders (VAE) and Generative Adversarial Networks (GAN) are commonly used, especially for high-dimensional inputs like images. Berend et al. \cite{berend2021distribution} propose the first distribution-guided coverage criterion, which integrates out-of-distribution (OOD) techniques to generate unseen test cases and provides a high level of assurance regarding the validity of identified faults in DNNs. In a study by Dola et al. \cite{dola2021distribution}, the validity of test cases generated by existing DNN test generation techniques is examined using VAE. By comparing the probability density estimates of a trained VAE model on data from the training distribution and OOD inputs, critical insights are obtained for validating test inputs generated by DNN test generation approaches. To generate realistic test cases that conform to requirements and reveal errors, Byun et al. \cite{byun2020manifold} employ a variant of Conditional Variational Autoencoder (CVAE) to capture a manifold that represents the feature distribution of the training data. Toledo et al. \cite{toledo2021distribution} introduces the first method called distribution-based falsification and verification (DFV), which utilizes environmental models to concentrate the falsification and verification of DNNs on meaningful regions of the input space. This method is designed to leverage the underlying distribution of data during the process of DNN falsification and verification. Huang et al. \cite{huang2022hierarchical} propose a hierarchical distribution-aware testing framework for DNNs. Their framework takes into account two levels of distribution: the feature level distribution, captured by generative models, and the pixel level distribution, which is represented by perceptual quality metrics. Although distribution aware testing can detect more meaningful faults for DNNs, which significantly contribute to the downstream repairing of DNNs, they still cannot provide the deterministic guarantee to the safety of DNNs.




\section{Conclusion}

Developing critical systems has always been challenging due to the potential harm caused by malfunctions, functional insufficiencies, or malicious attacks. The complexity is amplified when incorporating learning-enabled components, as the approaches taken by safety engineers who build the system often differ from those employed by AI/ML engineers who construct the components. Educating the general audience about AI safety concerns is essential for fostering active engagement in the ongoing discourse. However, to address the underlying engineering challenges, an interdisciplinary curriculum that bridges concepts from various fields such as AI/ML engineering and safety engineering can provide valuable insights and understanding.

We notice that there are two views on system safety in the broader community, the ``binary'' view and the ``probabilistic'' view, which present differing perspectives on how to approach safety assurance. Proponents of the binary view argue that safety is about clearly defining the system's capabilities and limitations, establishing a definitive ``safety boundary''. According to this view, we can confidently operate the system once we have a comprehensive understanding of this boundary. However, this viewpoint may hold primarily for traditional systems without AI components, where the system behavior is relatively simple and predictable.

The concept in this paper hints that we advocate the probabilistic view that safety for complex AI-enabled systems should be measured in terms of empirical probabilities\footnote{In statistics, empirical probability  refers to the probability of an event based on observed data or evidence. The empirical probability is also known as experimental probability because it is derived from actual experimentation or observation.}, as modern systems are becoming increasingly complex, with inherent uncertainties that make it difficult to determine the system's safety boundary precisely. In this perspective, the boundary itself may even appear blurred due to the non-deterministic behaviors exhibited by AI algorithms. Consequently, adherents of the probabilistic view assert that safety assurance should consider the likelihood of various outcomes and incorporate risk assessment and mitigation strategies to manage uncertainties effectively. 

 \bibliographystyle{elsarticle-num} 
 \bibliography{refs}

\end{document}

%% file: packages.tex
\usepackage[switch]{lineno}
\usepackage{float}
\usepackage{color}
\usepackage{graphicx}
\usepackage{comment}
\usepackage{wrapfig}
\usepackage{amsmath}
\usepackage{breakcites}

\usepackage{stix}

\usepackage{amsmath} 
\usepackage{centernot}
\usepackage{array}
\newcolumntype{P}[1]{>{\centering\arraybackslash}p{#1}}
\newcolumntype{M}[1]{>{\centering\arraybackslash}m{#1}}

\usepackage{amsfonts}
\usepackage{xcolor}
\usepackage{graphicx}

\usepackage[utf8]{inputenc}

\usepackage{todonotes}
\usepackage{amsfonts, amsmath, amssymb, bbm}
\usepackage{comment} 
\usepackage{xcolor}
\usepackage{tikz}
\usepackage{hyperref}
\usetikzlibrary{arrows,shapes,snakes,automata, calc, chains, matrix, positioning, scopes, arrows.meta}

\usepackage{tabularx}
\usepackage{subcaption} 
\usepackage{url}
\usepackage{xpatch}
\usepackage{lscape}
\usepackage[misc,geometry]{ifsym} 

\newcount\Comments  
\usepackage{color}
\definecolor{darkgreen}{rgb}{0,0.5,0}
\definecolor{purple}{rgb}{1,0,1}

%% file: macros.tex
\newcommand{\network}{{\cal N}}

\makeatletter
\newcommand{\xMapsto}[2][]{\ext@arrow 0599{\Mapstofill@}{#1}{#2}}
\def\Mapstofill@{\arrowfill@{\Mapstochar\Relbar}\Relbar\Rightarrow}
\makeatother

\usepackage{stackengine}

\newcount\Comments  
\definecolor{darkgreen}{rgb}{0,0.5,0}
\definecolor{purple}{rgb}{1,0,1}
\newcommand{\kibitz}[2]{\ifnum\Comments=1\textcolor{#1}{#2}\fi}

\newcommand{\xiaowei}[1]{{\color{blue}XH: #1}}

\newcommand{\red}{\color{red}}

\newcommand{\networks}{{\tt N}}
\newcommand{\inputdomain}{{\tt X}}
\newcommand{\outputdomain}{{\tt Y}}

\newcommand{\Comment}[1]{}